\documentclass[twoside,11pt,preprint]{article}

\usepackage{jmlr2e}

\usepackage{remreset}
\makeatletter
\@removefromreset{subsection}{section}

\makeatother

\usepackage{booktabs}

\usepackage{enumitem}

\newcommand{\new}[1]{{\color{red}#1}}
\renewcommand{\new}[1]{{#1}}
\newcommand{\citeN}[1]{\citeauthor{#1}~(\citeyear{#1})}

\usepackage{wrapfig}

\usepackage{titlesec}
\titleformat{\section}{\normalfont\large\bfseries}{}{0em}{\arabic{section}. }
\titleformat{\subsection}{\normalfont\large\bfseries}{}{0em}{Best Practice\ \arabic{subsection}: }
\newcommand{\bestprac}[1]{\subsection{#1}}

\usepackage{amsmath}
\usepackage{amssymb}

\newtheorem{defn}{Definition}
\newtheorem{ex}[defn]{Example}
\newtheorem{fact}[defn]{Fact}
\newtheorem{cor}[defn]{Corollary}

\usepackage{todonotes}

\jmlrheading{X}{20XX}{1-XX}{X/XX}{XX/XX}{linda19a}{Marius Lindauer and Frank Hutter}

\ShortHeadings{Best Practices for NAS}{Lindauer and Hutter}
\firstpageno{1}
\begin{document}

\title{Best Practices for Scientific Research\\ on Neural Architecture Search}

\author{\name Marius Lindauer \email lindauer@tnt.uni-hannover.de \\
       \addr Leibniz University of Hannover\\
       Hannover, 30167, Germany
       \AND
       \name Frank Hutter \email fh@cs.uni-freiburg.de\\
       \addr University of Freiburg \& Bosch Center for Artificial Intelligence\\
       Freiburg im Breisgau, 79110, Germany}

\editor{???}
% Heading arguments are {volume}{year}{pages}{date submitted}{date published}{paper id}{author-full-names}

\maketitle

\begin{abstract}
   Finding a well-performing architecture is often tedious for both depp learning practitioners and researchers, leading to tremendous interest in the automation of this task by means of neural architecture search (NAS).
   Although the community has made major strides in developing better NAS methods, the quality of scientific empirical evaluations in the young field of NAS is still lacking behind that of other areas of machine learning.
   To address this issue, we describe a set of possible issues and ways to avoid them, leading to the NAS best practices checklist available at \url{http://automl.org/nas_checklist.pdf}. 
\end{abstract}


%%%%%%%%%%%%%%%%%%%%%%%%%%%%%%%%%%%%%%%
\section{Introduction}
%%%%%%%%%%%%%%%%%%%%%%%%%%%%%%%%%%%%%%%

\begin{wrapfigure}{r}{0.5\linewidth}
\includegraphics[width=0.5\textwidth]{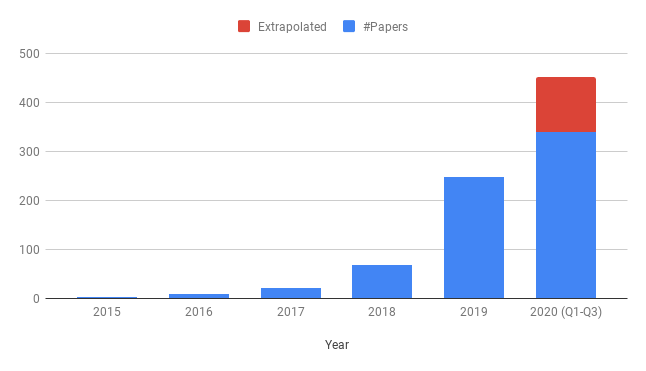}
    \caption{NAS papers per year based on the literature list on \url{automl.org}. \new{Extrapolation for 2020 based on the first 9 months of the year.}}
    \label{fig:nas_papers}
\end{wrapfigure}

Neural architecture search (NAS), the task of finding a well-performing architecture of a neural network for a given dataset, is currently one of the hottest topics in automated machine learning (AutoML; see \citet{automl_book} for an overview), with a seemingly exponential increase in the number of papers written on the subject (see Figure~\ref{fig:nas_papers})\footnote{The literature list on NAS at \url{www.automl.org} that this figure is based on is manually curated and lists all NAS papers we are aware of. The list contains both published papers and unreviewed arXiv papers.}. While many NAS methods are fascinating (see the survey article by \citet{elsken_neural_2018} for an overview of the main trends and a taxonomy of NAS methods), in this note we will not focus on these methods themselves, but on how to scientifically evaluate them and report one's findings.

Although NAS methods steadily improve, the quality of empirical evaluation in this field is still lagging behind compared to other areas in machine learning, AI and optimization.
\new{Over the last few years, we observed many times that results could not be reproduced because of one of the pitfalls mentioned throughout this paper. Although some might try to justify bad practices by the faster experimentation they allow, 
research on scientific methods shows that they rather increase confirmation bias~\citep{nickerson-rgp98a,fanelli-naos17a}. This will in turn slow down the overall research progress rather than speeding it up.}
We therefore propose best practices for empirical evaluations of NAS methods, which we believe will facilitate sustained and measurable progress in the field.

We note that discussions about reproducibility and empirical evaluations are currently taking place in several fields of AI. For example, Joelle Pineau's keynote at NeurIPS 2018\footnote{\texttt{https://videos.videoken.com/index.php/videos/}\newline \texttt{neurips-2018-invited-talk-on-reproducible-reusable-and-robust}\newline\texttt{-reinforcement-learning/}} showed how to improve empirical evaluations of reinforcement learning algorithms~\citep{henderson-aaai18a}, and several of her points carry over to NAS. For the NAS domain itself, \citeN{li-uai19a} also discuss reproducibility and simple baselines.

We resist the temptation to point to papers with flawed experiments, as no paper is perfect, including our own. However, to see examples for the pitfalls we mention, please randomly open five recent NAS papers, and you will very likely find examples for most of the pitfalls we mention and try to avoid.

We note that there are several recent papers that cast doubts on much of the work in the field of NAS~\citep{sciuto-arxiv19a,li-uai19a,xie-arxiv19a,Yang2020Frustrating}, and these have led to serious scepticism of outsiders concerning NAS. 
In this paper, we take into account all of the issues raised in those works, and several additional ones, in order to define a prescriptive set of best practices that will facilitate sustained progress in the field of NAS.

%%%%%%%%%%%%%%%%%%%%%%%%%%%%%%%%%%%%%%%
\section{Best Practices for Releasing Code}
%%%%%%%%%%%%%%%%%%%%%%%%%%%%%%%%%%%%%%%

Let's start with what is perhaps the most controversial set of best practices. This concerns reproducibility, a cornerstone of good science. As \citeN{buckheit_ws95a} put it: 

\emph{
“An article about computational science in a scientific publication is not the 
scholarship itself, it is merely the advertising of the scholarship. 
The actual scholarship is the complete software development environment and the complete set of instructions which generated the figures.”
}

Availability of code facilitates progress.
To facilitate fast progress in the field, it is important to be able to reproduce existing results. This helps studying and understanding existing methods, and to properly evaluate a new idea (see Section \ref{sec:comp_start}). 

Reproducing someone else's NAS experiments is often next to impossible without code.
The reproducibility crisis in machine learning has already shown how hard it is to reproduce each other's experiments without code in machine learning in general, but in NAS, this is further complicated by the fact that important settings are hidden both in the training pipeline (see Best Practice~\ref{sec:release_training}),  and in the NAS method itself (see Best Practice~\ref{sec:release_nas}). If the NAS-optimizer uses a neural network itself there is even more room for hidden choices. Therefore, we strongly advertise that each paper should come with a link to source code in order to facilitate reproducibility and sustained progress in the field.   

%%%%%%%%%%%%%%%%%%%%%%%%%%%%%%%%%%%%%%%
\bestprac{Release Code for the Training Pipeline(s) you use}\label{sec:release_training}
%%%%%%%%%%%%%%%%%%%%%%%%%%%%%%%%%%%%%%%

The training pipeline used is often far more important for achieving good performance than the precise neural architecture used~\citep{Yang2020Frustrating}. The training pipeline includes the specifics of the optimization and regularization methods used. For example, for image datasets, next to the choice of optimizer, number of training epochs, 
important choices include activation functions (e.g., Swish~\citep{elfwing-nn18a}), learning rate schedules (e.g., cosine annealing~\citep{loshchilov-iclr17a}), data augmentation (e.g. by CutOut~\citep{devries-arxiv17a}, MixUp~\citep{zhang-arxiv17a}
or Auto-Augment~\citep{cubuk-cvpr19a,lim-arxiv19a}),
auxiliary towers~\citep{zoph-cvpr18a}, the depth and width of the network, and regularization (e.g., by Dropout~\citep{srivastava-jmlr14a},  Shake-Shake~\citep{gastaldi-iclr17},  ScheduledDropPath~\citep{zoph-cvpr18a}, L1/L2 regularization, or decoupled weight decay~\citep{loshchilov-iclr19a}). For example, on CIFAR-10, using the search space from DARTS~\citep{liu-iclr19a}, the combination of CutOut, ScheduledDropPath, auxiliary towers, Auto-Augment, increasing the number of channels and increasing the number of epochs for training yielded a combined improvement of 3\% test accuracy, compared to less than 1\% for choosing the best neural architecture~\citep{Yang2020Frustrating}.

Therefore, the final performance results of paper A and paper B are \emph{incomparable} unless they use the same training pipeline. Releasing your training pipeline ensures that others can meaningfully compare against your results.
Especially the training pipeline for a dataset like CIFAR-10 should be trivial to make available, since this routinely consists of a single file relying only on open-source Tensorflow, Pytorch or MXNet code. Complex parallel training pipelines for larger datasets should also be easy to make available; even if there are some special dependencies that cannot be made available (e.g., a specialized framework for parallel training across many GPUs), availability of the main source code strongly facilitates reproducing results.

%%%%%%%%%%%%%%%%%%%%%%%%%%%%%%%%%%%%%%%
\bestprac{Release Code for Your NAS Method}\label{sec:release_nas}
%%%%%%%%%%%%%%%%%%%%%%%%%%%%%%%%%%%%%%%

While releasing the training pipeline allows researchers to fairly compare against your stated results, releasing the code for your NAS method allows others to also use it on new datasets. As an additional motivation next to following good scientific practice: papers with available source code tend to have far more impact and receive more citations than those without, because other researchers can build upon your code base.

%%%%%%%%%%%%%%%%%%%%%%%%%%%%%%%%%%%%%%%
\bestprac{Don't Wait Until You've Cleaned up the Code; That Time May Never Come}
%%%%%%%%%%%%%%%%%%%%%%%%%%%%%%%%%%%%%%%

We encourage anyone who can do so to simply put a copy of the code online as it was used, appropriately labelled as prototype research code, without using extra time to clean it up. This simply owes to the fact that, due to our busy lives as machine learners, the statement \emph{``The code will be available once I find the time to clean it up''} in practice all too often de facto (and without ill intent) translates to \emph{``The code will never be available''}. Of course, it is even better if you can release cleaned code in addition to the code dump we encourage. However, to make sure that you do release at all, please consider doing the code dump first. Indeed, we are pleased to observe that code releases are becoming far more common, partly due to the following fact and corollary. 

\begin{fact}
Reproducibility is ever more in the limelight.
\end{fact}
With the growing emphasis on reproducibility (e.g., as evidenced by ICML and NeurIPS now requesting authors to fill out Joelle Pineau's reproducibility checklist\footnote{\url{https://www.cs.mcgill.ca/~jpineau/ReproducibilityChecklist.pdf}} as part of the paper submission process), the trend at top machine learning venues is going towards authors having to justify cases in which code is not made available (and where the acceptance probability is reduced when no good reasons exist).  

\begin{cor}
A progressive policy for sharing code presents a competitive advantage in hiring for industrial research labs.
\end{cor}
Since top researchers want to publish in the top venues, and since this may become easier when sharing code, labs with a progressive policy for publishing code may soon have a competitive advantage in publishing at the top venues and thus in the global hunt for talent. We acknowledge that it is not always easy in industrial research environments to publish code, e.g., due to dependencies on proprietary components. However, there are by now many positive examples~\citep[e.g.,][]{pham-icml18a,liu-iclr19a,ying-icml19a} that demonstrate that sharing NAS code is possible for industrial players if it is made a priority. 

%%%%%%%%%%%%%%%%%%%%%%%%%%%%%%%%%%%%%%%
\section{Best Practices for Comparing NAS Methods}\label{sec:comp_start}
%%%%%%%%%%%%%%%%%%%%%%%%%%%%%%%%%%%%%%%

To ensure a fair comparison between NAS methods, but also to shed some light on the reasons that are responsible for a NAS method to perform well, we propose the following best practices.

%%%%%%%%%%%%%%%%%%%%%%%%%%%%%%%%%%%%%%%
\bestprac{Use the Same NAS Benchmarks, not Just the Same Datasets}\label{bp:nas_benchmark}
%%%%%%%%%%%%%%%%%%%%%%%%%%%%%%%%%%%%%%%

A very common way to compare NAS methods is a big table with the results different papers reported for a dataset such as \mbox{CIFAR-10}. However, we would like to emphasize that the numbers in these tables are often \emph{incomparable} due to the use of different search spaces and different optimization or regularization techniques (see also Best Practice~\ref{sec:release_training}). Rather, we propose the use of consistent \emph{NAS benchmarks}:

\begin{defn}[NAS Benchmark]\label{def:nas_bench}
A NAS benchmark consists of a dataset (with a pre-defined training-test split\footnote{If a validation set is required, this should be split off the training set; the test set should only be used for reporting the final performance. We do not request the validation set to be part of the definition of a NAS benchmark since different NAS methods may require validation sets of different sizes (e.g., only for hyperparameter optimization, or also for gradient-based architecture search).}), a search space\footnote{We note that the representation of the search space is sometimes also quite different. For example, it matters whether operations are in the nodes or on the edges.}, 
and available runnable code with pre-defined hyperparameters for training the architectures.
\end{defn}

\newpage
\noindent
We now give some good examples of such NAS benchmarks:

\begin{ex}
A prominent NAS benchmark is the publicly available search space and training pipeline of DARTS~\citep{liu-iclr19a}, evaluated on CIFAR-10 (with standard training/test split).\footnote{The only information that is unfortunately not available in their repository are the hyperparameter settings they used for their 100-epochs evaluation.}
\end{ex}

\new{
\citet{zela2020understanding} also defined an additional 12 smaller publicly available NAS benchmarks to demonstrate failure modes of DARTS. These are based on the search space of DARTS, but with fewer operators on the edges.

\begin{ex}
  Another prominent NAS benchmark is NAS-Bench-101 \citep{ying-icml19a},
  the first \emph{tabular} NAS benchmark. On top of a publicly available search space, training pipeline and dataset, it also provides pre-computed evaluations with that training pipeline for all possible cells in the search space. 
\end{ex}

Tabular NAS benchmarks speed up experimentation dramatically compared to NAS benchmarks that involve training neural networks for every evaluation, as they replace this expensive step by a table lookup. Further tabular NAS benchmarks along similar lines are NAS-Bench-1Shot1~\citep{zela2020nasbenchshot},  NAS-Bench-201~\citep{Dong2020NAS-Bench-201}, and
NAS-Bench-NLP~\citep{klyuchnikov2020nasbenchnlp}.\footnote{There is also a paper on NAS-Bench-ASR~\citep{nasbenchasr}, but the table for this tabular benchmark has not yet been released, and the code for generating it has not yet been open-sourced.}
While all tabular NAS benchmarks are inherently limited to small search spaces due to their requirement of exhaustively evaluating every architecture in the search space, \emph{surrogate} benchmarks scale to larger spaces: 

\begin{ex}
  NAS-Bench-301~\citep{siems-arxiv20a} is
  the first \emph{surrogate} NAS benchmark. On top of a publicly available large search space (the one of DARTS, with over $10^{18}$ architectures), training pipeline and dataset, it releases data for a subset of ~60k evaluated architectures in the search space, as well surrogate models that can be used to predict the performance of any architecture in the space. 
\end{ex}
}

We strongly believe that more such NAS benchmarks are needed for the community to make sustained and quantifiable progress (see also our note in Section \ref{sec:further_thoughts} concerning the need for new NAS benchmarks).

%%%%%%%%%%%%%%%%%%%%%%%%%%%%%%%%%%%%%%%
\bestprac{Run Ablation Studies}
%%%%%%%%%%%%%%%%%%%%%%%%%%%%%%%%%%%%%%%

NAS methods tend to have many moving pieces, some of which are more important than others. Also, unfortunately, some papers modify the NAS benchmark itself (e.g., the hyperparameters used for training the architecture; see Best Practices \ref{sec:release_training} and \ref{bp:nas_benchmark}), another component of the experimental pipeline, or various components of a NAS method
and leave it unclear which modification was most important to achieve their final results. 
%(on a dataset like \mbox{CIFAR-10}).
While NAS papers often get accepted based on these performance numbers, to strive for scientific insight, we should understand \emph{why} the final results are better than before. If a paper changes components other than the NAS method, then it is especially important to quantify the impact of these changes. 
Overall, as a community we also still lack thorough insights into what the most important aspects of NAS are, and only recently there is some progress on that front~\citep{sciuto-arxiv19a,li-uai19a,Yang2020Frustrating}.
Therefore, we recommend to run ablation analyses to study the importance of individual  components that affect performance, incl. any changes of the NAS benchmark and the NAS method itself. We would like to highlight the work by \citet{Yang2020Frustrating} as particularly valuable in this regards, \new{by showing that well engineered training protocols, the search space and macro architecture design substantially impact the overall performance.}

%%%%%%%%%%%%%%%%%%%%%%%%%%%%%%%%%%%%%%%
\bestprac{Use the Same Evaluation Protocol for the Methods Being Compared}\label{best:same_way}
%%%%%%%%%%%%%%%%%%%%%%%%%%%%%%%%%%%%%%%
So far, there is no single gold-standard on how to evaluate and compare NAS methods. In some cases, the outcome of a NAS run is only taken to be a single final architecture; in other cases, thousands of architectures are sampled and evaluated in order to select the best one. Of course, the latter is much less efficient, but it can lead to better performance. These different evaluation schemes are one of the reasons why results from different NAS papers are often incomparable.
Selecting the architecture with best performance on the \emph{test} set would of course lead to an optimistic estimate of performance, but selecting the best-performing architecture on a validation split is a perfectly reasonable building block of NAS algorithms; however, this step then becomes an integral part of the NAS method, and its runtime should be counted as part of the method (see also Best Practice \ref{best:end_to_end_time}).

%%%%%%%%%%%%%%%%%%%%%%%%%%%%%%%%%%%%%%%
\bestprac{Evaluate Performance as a Function of Compute Resources}\label{best:over_time}
%%%%%%%%%%%%%%%%%%%%%%%%%%%%%%%%%%%%%%%

While it is important to know the overall \new{compute resources} a NAS method required to obtain a result, it would be even more informative to report performance as a function of the required \new{compute resources}.\footnote{\new{Compute resources can be measured by different metrics, for example wallclock-time~\citep{coleman-sigops18a}, which will be hardware-dependent, or hardware-independent proxies, such as number of floating point operations (FLOPS), which is not always linearly correlated with the runtime~\citep{justus-ieee18a}, or in epochs (if these are roughly similarly expensive for different architectures and hyperparameter settings).}} This is possible in most cases since most NAS methods are anytime algorithms, and at each time point $t$ we can report the performance of the architecture that \emph{would} be returned if the search was terminated at $t$ (the so-called \emph{incumbent} architecture). 
This would also take into account that for some search spaces, it is trivial to obtain nearly the same performance as the optimal architecture, whereas for others this is quite hard. \new{We note that reporting the score of incumbents over \new{compute resources} comes at no extra cost when using tabular or surrogate NAS benchmarks.}

We also define two possible variants of NAS that differ w.r.t.\ how performance over \new{compute resources} should be measured:
\begin{defn}[Architecture identification variant of NAS] In this variant, only the \new{compute resources} for identifying and returning the final architecture counts. \label{def:arch_ident}
\end{defn}
In an offline evaluation using a private test set, in this variant, at each \new{compute resource} step $t$ we would plot the performance of the current incumbent architecture when trained with the final evaluation pipeline.

\begin{defn}[AutoML variant of NAS] In this variant, next to the \new{compute resources} for identifying the final architecture, we also count the \new{compute resources} to train the final architecture and return a model. \label{def:automl_nas}
\end{defn}
In an offline evaluation using a private test set, in this variant, at each step $t$, we would plot the performance of the current incumbent \emph{model} (not architecture), \emph{without retraining}. This reflects a true AutoML setting, where at the end of the compute budget one needs to be directly ready to make predictions.

Since NAS has not been used much in a full AutoML setting, the architecture identification variant is by far most widely used in the literature, but when all that we care about is a good model for a dataset at hand, the AutoML variant may be more suitable. 

%%%%%%%%%%%%%%%%%%%%%%%%%%%%%%%%%%%%%%%
\bestprac{Compare Against Random Sampling and Random Search}
%%%%%%%%%%%%%%%%%%%%%%%%%%%%%%%%%%%%%%%
As in other fields of machine learning, it is important for NAS research to compare against baselines. The simplest baselines are random sampling and random search~\new{\citep{bergstra-jmlr12a}}. Even though both of these rely on drawing uniform random samples from the architecture space, it is important to note that these are two different algorithms: 
\begin{defn}[Random sampling]
Random sampling draws a \emph{single} random sample from the architecture space and returns it. 
\end{defn}
The runtime of random sampling is therefore basically zero (wrt Definition~\ref{def:arch_ident}). Its expected performance is the average performance across all architectures in the search space.

\begin{defn}[Random search]
Random search draws random samples from the architecture space, evaluates them (with a criterion to be defined, such as a short training run or the full evaluation pipeline), and keeps track of the incumbent architecture with the best evaluation so far. At any given time, when stopped it returns this incumbent architecture. 
\end{defn}
Random search is an anytime procedure that should be run for the same amount of time as other approaches being compared to. To minimize confounding factors, it is useful to use the same evaluation criterion as the NAS method being compared to, both for random sampling and search.

Random sampling and random search are extremely simple procedures, but nevertheless many NAS papers avoid a comparison against these baselines. As \citeN{sciuto-arxiv19a} and \citet{xie-arxiv19a} show, random sampling can already yield strong performance in a well-designed search space, and \citeN{li-uai19a} show that random search can be very competitive.
Therefore, we recommend to compare against both of these baselines, to assess whether good performance is due to a well-designed search space (and training pipeline) or due to the NAS method.

%%%%%%%%%%%%%%%%%%%%%%%%%%%%%%%%%%%%%%%
\bestprac{Perform Multiple Runs with Different Seeds}
%%%%%%%%%%%%%%%%%%%%%%%%%%%%%%%%%%%%%%%
NAS methods are almost always stochastic. Therefore, re-running the same method on the same dataset does not necessarily lead to the same result~\citep{li-uai19a}. 
Additionally, some results can be quite hard to reproduce even if the source code is available. Sometimes, we observed that we needed several runs of the same code to reproduce the published results, indicating that the authors might have been lucky with the results reported in the paper. Therefore, we recommend that, if possible in terms of compute budgets, all methods should be repeated several times with different seeds and the authors should report mean and standard deviation (or median and quartiles if the noise is not symmetric) across the repetitions. \new{Ideally, one would control and report all used random seeds in a reproducible and simple sequence, e.g., random seeds from 1 to 10.} Besides improving the reproducibility of results, this will also provide new insights on the stochasticity of NAS methods in practice. For exact replicability, following~\cite{li-uai19a}, in either case we encourage the release of the exact seeds used for the NAS methods and final evaluation pipelines. 

%%%%%%%%%%%%%%%%%%%%%%%%%%%%%%%%%%%%%%%
\bestprac{Use Tabular or Surrogate Benchmarks If Possible}\label{sec:tabular}
%%%%%%%%%%%%%%%%%%%%%%%%%%%%%%%%%%%%%%%
We note that on standard NAS benchmarks, for most researchers, due to limited computational resources it will be impossible to satisfy the best practices in this section (especially ablation studies and performing several repeated runs). \new{Especially in such cases, we advocate running extensive evaluations on tabular NAS benchmarks, or on surrogate benchmarks as proposed by \citet{siems-arxiv20a} for NAS following the work of \citeauthor{eggensperger-aaai15a} (\citeyear{eggensperger-aaai15a,eggensperger-ml18a}).
We list available tabular and surrogate NAS benchmarks in Table~\ref{tab:tabular_and_surrogate_NAS}.}
\begin{table}[t!]
\new{
\small
    \centering
    \begin{tabular}{l l r r r}
        \toprule
        Name & Reference & \# arch  & supports & comments\\
         &  &  & one-shot? & \\        
        \midrule
        NAS-Bench-101 & \citet{ying-icml19a} & ~423k & no & constrained space\\
        NAS-Bench-1Shot1 & \citet{zela2020nasbenchshot} & ~6k -- ~364k & yes & 3 subspaces of NB-101\\
        NAS-Bench-201 & \citet{Dong2020NAS-Bench-201} & ~6k & yes & 3 datasets; learning curves\\
        NAS-HPO & \citet{klein-arxiv19z} & 62\,208 & no & 4 datasets; NAS + HPO\\
        NAS-Bench-NLP & \citet{klyuchnikov2020nasbenchnlp} & ~15k & yes & NLP\\
        NAS-Bench-301 & \citet{siems-arxiv20a} & $10^{18}$ & yes & surrogate benchmark\\
        \bottomrule
    \end{tabular}
    \caption{Overview over tabular and surrogate NAS benchmarks available so far.}
    \label{tab:tabular_and_surrogate_NAS}
}
\end{table}
These benchmarks allow even researchers without any GPU resources to perform systematic, comprehensive and reproducible NAS experiments by querying a table or a performance predictor instead of performing a costly optimization on special-purpose hardware. Importantly, by their very design, they also allow fair comparisons of different methods, without the many possible confounding factors of different training pipelines, hyperparameters, search spaces, and so on. 
We therefore advocate for running large-scale experiments on these tabular/surrogate benchmarks (studying the results of many repetitions, ablation studies, etc), and to complement these comprehensive experiments with additional small-scale experiments on real benchmarks.

We do note, however, that not all methods can be evaluated on tabular/surrogate benchmarks without retraining models (and therefore requiring substantial compute resources). While blackbox and multi-fidelity optimization methods \emph{can} be directly evaluated on these benchmarks only based on efficient lookups of performance, for weight sharing methods (e.g., DARTS~\citep{liu-iclr19a}) and weight inheritance methods (e.g., LEMONADE \citep{elsken2019efficient}) we cannot speed up the search process. We \emph{can}, however, still speed up the  evaluation of the selected architectures (by just looking up their performance) and \new{can likewise plot the performance of the incumbent architecture as a function of the compute resources, as proposed by~\citet{zela2020nasbenchshot}. }

Since the search phase of weight sharing and weight inheritance methods still needs to be run fully even for tabular/surrogate benchmarks, authors should not be expected to execute many runs of these methods (even when using tabular/surrogate benchmarks).

%%%%%%%%%%%%%%%%%%%%%%%%%%%%%%%%%%%%%%%
\bestprac{Control Confounding Factors}\label{bp:confounding}
%%%%%%%%%%%%%%%%%%%%%%%%%%%%%%%%%%%%%%%

Even when different papers use the same NAS benchmark, the performance results they report are still often incomparable due to various other confounding factors, such as 
different hardware, different runtimes used, 
and even different versions of DL libraries. 
All these details can substantially impact the results, and we therefore recommend that such confounding factors should be controlled as much as possible.
One convenient way of achieving unbiased apples-to-apples comparisons would be to perform experiments using an open-source library of NAS methods that allows running all methods without any confounding factors; see also our note in Section \ref{sec:further_thoughts} concerning such a library.

%%%%%%%%%%%%%%%%%%%%%%%%%%%%%%%%%%%%%%%
\section{Best Practices for Reporting Important Details}
%%%%%%%%%%%%%%%%%%%%%%%%%%%%%%%%%%%%%%%

To ensure reproducibility, it is important to report all the little, but important details that were responsible for the performance achieved by a NAS method. In the following, we propose best practices for details that are often missing but crucial in our experience.

%%%%%%%%%%%%%%%%%%%%%%%%%%%%%%%%%%%%%%%
\bestprac{Report the Use of Hyperparameter Optimization}
%%%%%%%%%%%%%%%%%%%%%%%%%%%%%%%%%%%%%%%
A particularly important detail is the hyperparameter optimization approach used. While the hyperparameters of the final evaluation pipeline are part of the NAS benchmark used (see Definition \ref{def:nas_bench}) and thus should not be changed without good reason and appropriate emphasis in the reporting of results, every NAS method also has its own hyperparameters. It is well known that these hyperparameters can influence results substantially -- e.g., for DARTS~\citep{liu-iclr19a}, they can make the difference between state-of-the-art performance and converging to degenerate architectures with very poor performance~\citep{zela2020understanding}.
Therefore, firstly (and connected to Best Practices~\ref{sec:release_training}, \ref{sec:release_nas}, \ref{bp:nas_benchmark} and \ref{bp:confounding}), the used hyperparameter setting is an important experimental detail that should be reported. Secondly, how this setting was obtained is important for applying a NAS method to a new dataset (which may require a different setting). 
Finally, when facing a new dataset, the time required for hyperparameter optimization should be considered as part of a NAS method's runtime. More than once we have heard statements like \emph{``Of course, NAS method X does not work out of the box for a new dataset, you first need to tune its hyperparameters''}, and we note that this should ring an alarm bell since this just replaces manual architecture engineering by manual hyperparameter optimization of the NAS method. Also, AutoML, by its very definition, needs to be robust; therefore, when viewed from an AutoML point of view, the hyperparameter optimization strategy in essence becomes part of the NAS method and ought to count as part of its runtime. Relatedly, we would also like to remark that statements like \emph{``We only applied a limited amount of hyperparameter optimization''} or \emph{``We slightly tuned the hyperparameters''} are too vague and not useful for reproducing results in a scientific way.

%%%%%%%%%%%%%%%%%%%%%%%%%%%%%%%%%%%%%%%
\bestprac{Report End-to-End Resources Required for the Entire NAS Method}\label{best:end_to_end_time}
%%%%%%%%%%%%%%%%%%%%%%%%%%%%%%%%%%%%%%%

Related to Best Practice~\ref{best:over_time}, we note that the \new{compute resources} required for a NAS method should be measured in an end-to-end fashion.
This is 
particularly important if different NAS methods run differently (see also Best Practice~\ref{best:same_way}). In particular, some NAS methods propose multiple potential architectures after a first phase and then select the final architecture among these in a validation phase. In such a case, in addition to reporting the \new{compute resources} for the individual phases, the \new{compute resources} required for the validation phase have to be counted as part of the overall \new{resources} used for the NAS method. 

\begin{ex}If a NAS method performs $k$ parallel search runs of time $T_{search}$ and selects the best of the resulting $k$ architectures in a validation phase that takes time $T_{valid}$  for each of the $k$ architectures, and the final architecture takes time $T_{final}$ to train, then the time requirement of the NAS method should be reported as $k \cdot (T_{search} + T_{valid})$ in the architecture identification variant (Definition~\ref{def:arch_ident}), and as $k \cdot (T_{search} + T_{valid}) + T_{final}$ in the AutoML variant (Definition~\ref{def:automl_nas}).
\end{ex}

%%%%%%%%%%%%%%%%%%%%%%%%%%%%%%%%%%%%%%%
\bestprac{Report All the Details of Your Experimental Setup}
%%%%%%%%%%%%%%%%%%%%%%%%%%%%%%%%%%%%%%%
These days, one of the main foci in NAS is to obtain good architectures faster. Therefore, results typically include the achieved accuracy (or similar metrics) and the time used to achieve these results. However, to assess and reproduce such results, it is important to know the hardware used (type of GPU/TPU, etc) and also the deep learning libraries and their versions.\footnote{Deep Learning libraries, such as tensorflow, pytorch and co are getting more efficient over time, but which version was actually used is unfortunately only reported rarely.} If method A needed twice as much time as method B, but method A was evaluated on an old GPU and method B on a recent one, the difference in GPU may explain the entire difference in speed. Overall, we recommend to report all the details required to reproduce results---all top machine learning conferences allow for a long appendix, such that space is never a  reason to omit these details.

For anyone wanting to publish code (also see Best Practice~\ref{sec:release_nas}), in order to avoid forgetting to include any software dependencies that are publicly available, a possibility to consider would be to publish a container, e.g., based on Docker or Singularity.

\titleformat{\subsection}{\normalfont\large\bfseries}{}{0em}{}

%%%%%%%%%%%%%%%%%%%%%%%%%%%%%%%%%%%%%%%
\section{Further Ways Forward For the Community}\label{sec:further_thoughts}
%%%%%%%%%%%%%%%%%%%%%%%%%%%%%%%%%%%%%%%

Besides striving for the best practices listed above, we identified two structural problems in the NAS community. We believe that addressing these will facilitate NAS research.

%%%%%%%%%%%%%%%%%%%%%%%%%%%%%%%%%%%%%%%
\subsection*{The Need for Proper NAS Benchmarks}
%%%%%%%%%%%%%%%%%%%%%%%%%%%%%%%%%%%%%%%

The seminal paper by \citet{zoph-iclr17a} 
used the \mbox{CIFAR-10} and PTB datasets for its empirical evaluation, and more than 300 NAS papers later, these datasets still dominate in empirical evaluations. 
While this is nice in terms of comparing methods on standardized datasets, it also involves a big risk of overfitting NAS to them. %
From a meta learning point of view, we are testing on our training set of two samples -- which is obviously not a good basis for developing methods that apply in general.

We do not argue for abandoning these datasets, but we do argue for the creation of a larger, standardized suite of well-defined \emph{NAS benchmarks}. Recall from Definition \ref{def:nas_bench} that such a NAS benchmark includes not only a dataset, but also a search space and a training pipeline with fully available source code and known hyperparameters. For CIFAR-10 and PTB, we do have access to proper NAS benchmarks, based on the search spaces and source code from the DARTS paper~\citep{liu-iclr19a}, and it would be very helpful for the community to have more of these. \new{Access to a set of interesting NAS benchmarks would allow the community to satisfy Best Practice \ref{bp:nas_benchmark} (comparing on the same benchmarks) by construction, and it would also make Best Practice \ref{bp:confounding} (controlling confounding factors) much easier to follow, since it would allow developers of new NAS methods to compare these under (nearly) the same conditions. }

We note that application papers in NAS have already started tackling non-standard applications, such as image restoration~\citep{pmlr-v80-suganuma18a}, semantic segmentation~\citep{chen-nips18a, Nekrasov_semseg,liu-arxiv19a}, disparity estimation~\citep{saikia_arXiv19}, machine translation \citep{so_transfomrer}, 
reinforcement learning~\citep{runge2019learning}, and GANs~\citep{AutoGAN}.
However, to the best of our knowledge, none of these papers make a clean new NAS benchmark available (as defined above) to complement \mbox{CIFAR-10} and PTB.\footnote{\new{However, we do note that while this paper was under review, new works introduced tabular NAS benchmarks for natural language processing (NLP,~\cite{klyuchnikov2020nasbenchnlp}) and automatic speech recognition (ASR,~\cite{nasbenchasr}).}} 

We therefore encourage researchers who work on exciting applications of NAS to create new NAS benchmarks based on their applications. In fact, we believe that at this point of time, a paper that simply evaluates \emph{existing} NAS methods on a new exciting application and makes available a new fully reproducible NAS benchmark based on this would have a more lasting positive impact on the development of the NAS community than a paper introducing a slightly improved NAS method.

\new{In the long run, we envision a library of NAS benchmarks that is (i) diverse with respect to (a) difficulty, (b) search space properties, (c) datasets and applications, (d) training pipelines; and (e) a mix of real, tabular, and surrogate benchmarks; and that has (ii) a standardized API which allows developers of NAS methods to easily benchmark their ideas on them. Although this might be an ambitious goal in view of the young field, other meta-domains showed that it is in fact feasible and worthwhile~\citep{eggensperger-bo13,hutter-lion14,bischl-aij16a}. First steps towards such libraries are already taken by the series of NAS-Bench papers~\citep{ying-icml19a,zela2020nasbenchshot,Dong2020NAS-Bench-201,siems-arxiv20a,klyuchnikov2020nasbenchnlp}. 

\subsubsection*{Properties of a NAS Benchmark}
To obtain an interesting set of NAS benchmarks, we propose to aim for diversity in the following properties for NAS search spaces:
\begin{description}
    \item[Difficulty] How hard is it to achieve a good performance on the given benchmark? Many of the current NAS benchmarks are relatively easy in the sense that even random search performs very well (often within 0.5\% of optimal), and we would encourage also including some benchmarks where the best architectures perform substantially better than the ones found by random search.
    \item[Expressive Power] How many architectures can be designed based on the given search space? These days most efficient NAS methods search on fairly limited search spaces (like the typical cell search spaces), and it would be useful to also include much larger search spaces.
    \item[Complexity] How complex is the search space? Possible examples for quantifying complexity include the number of allowed operators, the potential complexity of pathways through the network, the density of possible connections, etc.
    \item[Novelty] Does the search space allow to find novel architectures, or does it only include known types of architectures? For new tasks, we are sometimes looking for completely novel architectures, since no well performing architectures are yet known for the task. But on other tasks, we might already know which architectures are likely to work well and are only interested in finding the best architecture in this limited space.
    \item[Achievable performance] Since practical applications of NAS aim for obtaining new state-of-the-art performance it is useful for NAS benchmarks to include networks that yield competitive performance (rather than only optimizing architectures that are clearly dominated). However, NAS benchmarks that have proven useful to the community should not be disregarded once other types of networks achieve better performance, but should still be preserved to warrant comparability of results over the years and also provide a potentially cheaper experimentation environment than the newest (and likely most expensive) benchmarks.
    \end{description}

Likewise, we strongly encourage a library of NAS benchmarks to include a set of diverse applications, datasets and training pipelines.

\subsubsection*{Practical Recommendations for Tabular and Surrogate NAS Benchmarks}
Since several new tabular NAS benchmarks are being introduced these days, we would also like to give some recommendations on how to create a good tabular NAS benchmark:
\begin{enumerate}
    \item Release the table in an easy-to-access form. Papers reporting on tabular NAS benchmarks for which the table is not available are not very useful for the community.
    \item Release code. While a tabular benchmark can be useful in itself, it is far more useful if the code for generating it is also released (again, along with hyperparameters and, optimally, seeds). If code is not available, the benchmark cannot be used to evaluate NAS methods based on weight sharing or weight inheritance.
    \item If possible, keep track of, and release other metrics being computed, such as the number of weights, flops, etc.
\end{enumerate}

Likewise, for building surrogate NAS benchmarks, as discussed by \citet[][Appendix F]{siems-arxiv20a}, special consideration should be paid to data collection, construction of several surrogate models, verification and validation, version numbering, and release of training data \& source code. 
}

\subsection*{The Need for an Open-Source Library of NAS Methods}

In addition to a well-defined benchmark suite, 
there is also a need for an open-source library of NAS methods that allows for (i) a common interface to NAS methods, (ii) the control of confounding factors, (iii) fair and easy-to-run comparisons of different NAS methods on several benchmarks and (iv) an assessment of how important each component of a NAS method is.
\new{Although several well-engineered and flexible AutoDL packages were recently proposed~\citep{jin-sigkdd19a,omalley2019kerastuner,erickson-arxiv20a,zimmer-arxvi20a}, their main goal is ease of use, rather than allowing clean comparisons of several NAS methods.}
Libraries of methods have had a very positive impact on other fields (e.g., RLlib~\citep{liang-icml18}, Tensorforce~\citep{tensorforce} or OpenAI~\citep{baselines}), and we expect a similarly positive impact for the field of NAS.
\new{The DeepArchitect library~\citep{negrinho2017deeparchitect} already proposed a unified search space definition, and several recent projects are making further progress along these lines~\citep{shah-archai20,ning-arxiv20a,naslib}.}
Optimally, an open-source library would allow researchers to implement their algorithms easily and in a search-space-independent fashion to directly follow most of the best practices given here by minimizing confounding factors. \new{If the library allows unified access to a broad collection of NAS benchmarks, and if it allows the comparison of different methods without confounding factors, work carried out on top of it will satisfy many of the best practices described here by construction.} 
Besides facilitating NAS research, such a library could also have a great impact by giving interested users of NAS access to the best current NAS approaches. 

%%%%%%%%%%%%%%%%%%%%%%%%%%%%%%%%%%%%%%%
\section{Conclusion}
%%%%%%%%%%%%%%%%%%%%%%%%%%%%%%%%%%%%%%%

We proposed $14$ best practices for scientific research on neural architecture search (NAS) methods. We believe that gradually striving for them as guidelines will increase the scientific rigor of NAS papers and help the community to make sustained progress on this key problem.

Similar to Joelle Pineau's reproducibility checklist, we have compiled the best practices for NAS research described here into a checklist for authors and reviewers alike. We hope that this checklist will help to easily assess the state of a paper. 
This checklist is available at the following URL: \url{http://automl.org/NAS_checklist.pdf}.

%%%%%%%%%%%%%%%%%%%%%%%%%%%%%%%%%%%%%%%
\section*{Acknowledgement}
%%%%%%%%%%%%%%%%%%%%%%%%%%%%%%%%%%%%%%%
We thank Thomas Elsken, Arber Zela, Katharina Eggensperger, Matthias Feurer, as well as the anonymous reviewers, for comments on an earlier draft of this note.

%%%%%%%%%%%%%%%%%%%%%%%%%%%%%%%%%%%%%%%
\bibliography{best_practices.bib}
%%%%%%%%%%%%%%%%%%%%%%%%%%%%%%%%%%%%%%%

\end{document}